\documentclass[preprint,12pt]{elsarticle}

\usepackage{amssymb}
\usepackage{amsmath}
\usepackage{latexsym}
\usepackage{amssymb}
\usepackage{amsmath}
\usepackage{amsthm}
\usepackage{booktabs}
\usepackage{enumitem}
\usepackage{graphicx}
\usepackage{color}

\usepackage{colortbl}
\usepackage[dvipsnames]{xcolor}
\usepackage{multirow}
\usepackage{subcaption}
\definecolor{headerbg}{RGB}{0, 51, 102} % Dark blue background
\definecolor{headerfg}{RGB}{255, 255, 255}
\definecolor{lightblue}{rgb}{0.88, 0.95, 1.0} 
\usepackage{textcomp}
\usepackage{xcolor}

\journal{Pattern Recognition}
\date{}
\begin{document}

\begin{frontmatter}

\title{From Video to EEG: Adapting Joint Embedding
Predictive Architecture to Uncover Visual Concepts in
Brain Signal Analysis} %% Article title

% Manually formatted author block
\author{
    Amirabbas Hojjati$^{\ast\dagger}$,
    Lu Li$^{\P\dagger}$,
    Ibrahim Hameed$^{1}$,
    Anis Yazidi$^{\star\S}$,
    Pedro G. Lind$^{\star\ddagger}$,
    Rabindra Khadka$^{\star}$\corref{cor1}}

% Corresponding author note
\cortext[cor1]{Corresponding author. Email: \texttt{rabindra@oslomet.no}}

% Affiliations + contribution note as footnote
\fntext[fn1]{
$^{\dagger}$ Equal contribution.
$^{\P}$ Sun Yat-sen University, China. 
$^{1}$ Norwegian University of Life Sciences, Norway. 
$^{\S}$ University of Oslo, Norway. 
$^{\ddagger}$ Simula Research Laboratory, Norway. 
$^{\ast}$ Timely, Norway. 
$^{\star}$ Oslo Metropolitan University, Norway. 
}

% %% Authors
% \author[E]{Amir Hojjati\footnotemark[1]}
% \author[A]{Lu Li\footnotemark[1]}
% \author[B]{Ibrahim Hameed}
% \author[C,F]{Anis Yazidi}
% \author[D,F]{Pedro G. Lind}
% \author[D]{Rabindra Khadka\corref{cor1}}

% \cortext[cor1]{Corresponding author. Email: \texttt{rabindra@oslomet.no}}

% %% Affiliations

% \affiliation[A]{organization={Sun Yat-sen University},
%             % addressline={},
%             % city={},
%             % postcode={},
%             % state={},
%             country={China}}

% \affiliation[B]{organization={Norwegian University of Life Sciences},
%             % addressline={},
%             % city={},
%             % postcode={},
%             % state={},
%             country={Norway}}

% \affiliation[C]{organization={University of Oslo},
%             % addressline={},
%             % city={},
%             % postcode={},
%             % state={},
%             country={Norway}}

% \affiliation[D]{organization={Simula Research Laboratory},
%             % addressline={},
%             % city={},
%             % postcode={},
%             % state={},
%             country={Norway}}
            
% \affiliation[E]{organization={Timely},% Department and Organization
%             % addressline={},
%             % city={},
%             % postcode={},
%             % state={},
%             country={Norway}}
            
% \affiliation[F]{organization={Oslo Metropolitan University},
%             % addressline={},
%             % city={},
%             % postcode={},
%             % state={},
%             country={Norway}}

%% Abstract

\begin{abstract}
EEG signals capture brain activity with high temporal but low spatial resolution, supporting applications such as neurological diagnosis, cognitive monitoring, and brain-computer interfaces. However, effective analysis remains challenging due to limited labeled data, high dimensionality, and the lack of scalable models that fully capture spatiotemporal dependencies. Existing self-supervised learning (SSL) methods often focus on either spatial or temporal features in isolation, leading to suboptimal representations. To this end, we propose EEG-VJEPA, a novel adaptation of the Video Joint Embedding Predictive Architecture (V-JEPA) for EEG classification. By treating EEG as video-like sequences, EEG-VJEPA learns semantically meaningful spatiotemporal representations using joint embeddings and adaptive masking. To our knowledge, this is the first work that exploits V-JEPA for EEG classification and explores the visual concepts learned by the model. EEG-VJEPA achieves state-of-the-art performance on the publicly available Temple University Hospital (TUH) Abnormal EEG dataset, outperforming both self-supervised and fully supervised baselines. Likewise, we demonstrate the model’s good generalization ability on an independent, smaller clinical dataset from the General Hospital of Thessaloniki, involving dementia classification. Beyond classification, EEG-VJEPA captures physiologically meaningful spatiotemporal patterns, offering interpretable embeddings useful for human-AI collaboration in clinical workflows. These results highlight its potential as a scalable and trustworthy framework for real-world EEG analysis.
\end{abstract}


% \begin{highlights}
% \end{highlights}

%% Keywords

\end{frontmatter}


\section{Introduction}

Electroencephalography (EEG) is a non-invasive and cost-effective technique for capturing rhythmic brain activity, widely used in clinical neurology to monitor conditions such as epilepsy, encephalopathy, and cognitive disorders~\cite{michel2012towards, mcfarland2017eeg}. With the growing need for scalable diagnostic tools, integrating artificial intelligence (AI) into EEG interpretation can enhance early diagnosis and reduce clinical workload.  However, most ML models rely on supervised learning, which demands large amounts of labeled data; a resource often difficult and costly to curate for EEG~\cite{rafiei2022self,jiang2021self,kostas2021bendr,krishnan2022self}. In response, self-supervised learning (SSL) has emerged as a promising alternative to leverage large-scale unlabeled EEG data~\cite{ye2024self,dong2025self,zhang2025psss}. Recent advances in SSL for time series and EEG~\cite{kostas2021bendr,jiang2021self,mohsenvand2020contrastive} have adapted paradigms originally developed for vision and language domains, such as contrastive learning and masked signal modeling~\cite{chen2020simple,zbontar2021barlow,he2022masked}. Yet, existing approaches often struggle to capture the intricate spatiotemporal characteristics of multi-channel EEG data. Invariance-based SSL methods~\cite{zbontar2021barlow,chen2020simple,grill2020bootstrap} require careful design of augmentations and inductive biases, while generative approaches\cite{he2022masked,feichtenhofer2022masked} may fail to learn semantically meaningful representations relevant for downstream tasks~\cite{assran2022hidden,li2022multi}.

%SSL approaches include invariance-based methods, which optimize encoders to produce consistent embeddings across augmented views, and generative methods, which focus on reconstructing corrupted signals~\cite{chen2020simple,grill2020bootstrap,zbontar2021barlow,he2022masked,feichtenhofer2022masked}. While invariance-based methods are limited by strong inductive biases and the challenge of selecting augmentations, generative methods reduce these biases but produce low-level semantic representations~\cite{mohsenvand2020contrastive,assran2022hidden,tian2020self,li2022multi}.
In this work, we introduce \textbf{EEG-VJEPA}, a novel self-supervised framework that extends JEPA to model multi-channel EEG as spatiotemporally masked sequences, analogous to video frames. Using a Vision Transformer (ViT) backbone, our model captures both spatial and temporal dependencies. EEG-VJEPA achieves the state-of-the-art results in abnormal EEG classification on the TUAB dataset~\cite{harati2014tuh}, outperforming EEG2REP~\cite{foumani2024eeg2rep}, LaBraM~\cite{jiang2024large}, and contrastive learning model~\cite{hojjati2023multi} by 6.4\%, 4\%, and 2.45\%, respectively.

\begin{figure*}[t]
    \centering
    \includegraphics[width=\textwidth,]{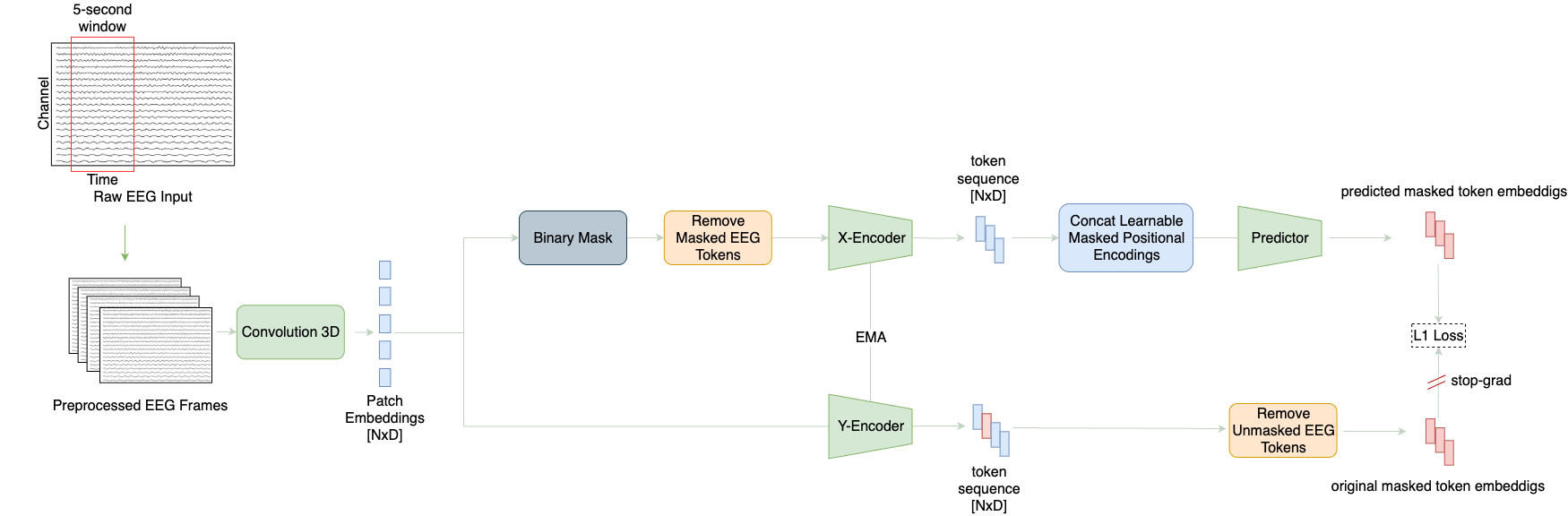}
    \caption{EEG-VJEPA. EEG signals are transformed into 3D shapes using a sliding window technique. The input passes through 3D convolution to produce patch embeddings with spatiotemporal features. The sequence of flattened tokens is masked before feeding into the X-encoder. Learnable masked tokens with positional embedding are added to the output of the X-encoder. The predictor network outputs embedding vectors for each mask token. The Y-encoder processes the input signal without masking to output the ground truth for the predictor. L1 loss is applied, as recommended in \protect\cite{bardes2024revisiting} for stability, minimizing the distance between the output of the predictor and the Y-encoder. The parameters of the Y-encoder are updated using the exponential moving average (EMA) of the X-encoder weights. }
\vspace{1em}
    \label{pipeline}
\end{figure*}

More importantly, EEG-VJEPA exhibits several properties that are desirable for real-world clinical adoption. It produces embeddings that align with physiologically interpretable EEG patterns, shows robustness to intersubject variability, and identifies relevant spatiotemporal signal regions. These findings position EEG-VJEPA not only as a good-performing model but as a foundation model candidate for downstream clinical applications, including disease triage, risk stratification, and decision support.

We highlight the following contributions of our work:

\begin{enumerate}
    \item We present \textbf{EEG-VJEPA}, a novel self-supervised model based on the joint embedding predictive architecture that incorporates the spatial and temporal attributes of EEG signals. We treat EEG signals as video-like sequences, enabling the application of V-JEPA's spatiotemporal masking strategy.
    
    \item We demonstrate that self-supervised learning with the VJEPA strategy yields substantial performance gains over EEG2REP~\cite{foumani2024eeg2rep}, LaBraM~\cite{jiang2024large}, and contrastive learning-based models~\cite{hojjati2023multi} on the clinically relevant TUAB dataset, driven by the rich spatiotemporal representations learned in the latent space. Furthermore, we validate the generalization ability of EEG-VJEPA on an independent, smaller clinical dataset from the General Hospital of Thessaloniki. 
    
    \item Through extensive ablation studies and downstream evaluations, we explore the impact of various hyperparameters on model performance. We find that certain settings, such as patch size and augmentation strategies, can significantly enhance the model's performance.

    \item Finally, we observe that EEG-VJEPA effectively learns the spatiotemporal dynamics of EEG signals, localizing meaningful regions of interest, without access to labels. We show that the embeddings learned by the model capture physiologically and clinically relevant features.    
    
   % We present visualizations of the learned representations, analyzing variations across different aspects such as age and classes. We also present attention rollout maps to show attention propagation across the channel and time dimensions, highlighting the spatiotemporal dynamics of the model as it processes EEG data.

\end{enumerate}

The remainder of the paper is structured as follows: Section 2 provides an overview of the SSL literature. In Section 3, we present the methodology of our work. Section 4 describes the experimental settings, highlighting the implementation details and results. In Section 5, we discuss our findings and implications. Finally, in section 6, we conclude the paper by discussing its limitations and outlining potential directions for future research.
%%%%%%%%%%%%%%%%%%%%%%%%%%%%%%%%%%%%%%%%%%%%%%%%%%%%%%%%%%%%%%%%%%%%%%%%

\section{Related Work}

 Substantial progress has been made in EEG representation learning, with advances in feature extraction, pre-training, predictive modeling, and downstream task performance. Early works transformed time-series EEG data into frequency domains and combined statistical and spectral features for classification using models such as SVMs~\cite{sun2019multi}. Other studies leveraged joint representations from time and frequency domains via convolutional layers~\cite{liu2019epileptic} or multi-view approaches such as wavelet transforms and attention-based gated recurrent units~\cite{tang2020seizure}. Incorporating brain connectivity measures like phase lag index and power spectral density has also enhanced predictive modeling~\cite{chen2022fusing}.

Self-supervised learning (SSL) methods have shown promise in addressing the scarcity of labeled EEG data. Multi-view SSL approaches have fused diverse representations, such as time, frequency, and spatial features, for tasks like motor imagery classification and seizure prediction~\cite{xu2020recognition,liu2019epileptic}. A work by~\cite{hojjati2023multi} developed a self-supervised approach for EEG representation learning using multi-view contrastive learning. The method utilized both time and frequency domain views of EEG data with data augmentation. \cite{wan2023eegformer} introduced EEGformer, a model combining convolutional and Transformer layers to capture temporal and regional EEG features. These methods demonstrated strong performance in emotion recognition and seizure detection. Recent works also highlight the potential of large-scale pre-training for EEG. Chen et al.~\cite{chen2024eegformer} developed a 1.7TB EEG foundation model using a vector-quantized Transformer, offering interpretable representations with strong transfer learning capabilities. The Joint Embedding Predictive Architecture (JEPA), a self-supervised framework for learning semantic representations by predicting missing information in the latent space without data augmentation, was introduced by~\cite{assran2023self}. This approach has been adapted for various data types, including EEG signals. For instance, Guetschel et al. proposed Signal-JEPA which incorporates a domain-specific spatial block masking strategy to enhance cross-dataset transfer in EEG signal processing~\cite{guetschel2024s}. However, these approaches often fail to capture the long-range temporal context critical for EEG analysis. Inspired by V-JEPA~\cite{bardes2024revisiting}, we extend this architecture to EEG by treating signals as video-like sequences.

\begin{table}[ht]
 \caption{Summary of the main hyper-parameters tuned for pre-training. The highlighted values of the parameters indicate those that provide the best pre-trained EEG-VJEPA model.}
    \centering
    \arrayrulecolor{headerbg} % Set border color
    \resizebox{\textwidth}{!}{%
    \begin{tabular}{|c|c|c|c|}
        \hline
        \rowcolor{headerbg} % Apply background color to header
        \textcolor{headerfg}{\textbf{Hyperparameters}} & \textcolor{headerfg}{\textbf{ViT-B}} & \textcolor{headerfg}{\textbf{ViT-M}} & \textcolor{headerfg}{\textbf{ViT-S}} \\
        \hline

        \multirow{1}{*}{batch \textunderscore size} & 10,30,40 & 6,\colorbox{GreenYellow}{10},30 & 4,5,10,30 \\
        \hline
        predictor \textunderscore depth & 12 &\colorbox{GreenYellow}{6}  & 4 \\
        \hline
        sampling \textunderscore rate & 1,2 & 2,\colorbox{GreenYellow}{3} & 1,2,3 \\
        \hline
        frames  & 16,32,64 & 16, \colorbox{GreenYellow}{32} & 16,32 \\
        \hline
        clips  & 1,3,6 & \colorbox{GreenYellow}{1},3 & 1,3,6 \\
        \hline
        \multirow{7}{*}{tubelet \textunderscore size ($h \times w \times t$)} & $2 \times 30 \times 2$ & \colorbox{GreenYellow}{$4 \times 30 \times 4 $} & $2 \times 30 \times 2$ \\
        \cline{2-4}
        & $2 \times 50 \times 2$ & $2 \times 30 \times 2$ & $4 \times 30 \times 4$ \\
        \cline{2-4}
        & $2 \times 15 \times 2$ & -  & - \\
        \cline{2-4}
        & $4 \times 30 \times 2$ & - & - \\
        \cline{2-4}
        & $2 \times 30 \times 8$ & - & - \\
        \cline{2-4}
        & $2 \times 30 \times 4$ & - & - \\
        \cline{2-4}
        & $4 \times 30 \times 4$ & -  & - \\
        \hline
        \multirow{2}{*}{augmentation} & spatial, noise, & \multirow{2}{*}{\colorbox{GreenYellow}{spatial}}& \multirow{2}{*}{flip, spatial} \\
        &  scale, flip & & \\
        \hline
        random \textunderscore resize \textunderscore aspect \textunderscore ratio & (0.75,1.35) & \colorbox{GreenYellow}{(0.75,1-35)} & (0.75,1.35) \\
        \cline{2-4}
        & - & (0.75,1.0) & - \\
        \hline
        \multirow{2}{*}{random \textunderscore resize \textunderscore scale} & (0.3,1.0) & \colorbox{GreenYellow}{(0.3,1.0)} & (0.3,1.0) \\
        \cline{2-4}
        & - & (0.75,1.0) & - \\
        \hline
    \end{tabular}%
    }
   
\label{hyper_params}.
%\vspace{-1em}
\end{table}

%%%%%%%%%%%%%%%%%%%%%%%%%%%%%%%%%%%%%%%%%%%%%%%%%%%%%%%%%%%%%%%%%%%%%%%%

\section{Methodology}

\noindent\textbf{Joint-Embedding Predictive Architecture:}
%\hspace{0.25cm}
The architecture consists of two encoders (X-encoder and Y-encoder) and a predictor network, as illustrated in Figure~\ref{pipeline}. The encoders are parameterized as Vision Transformers (ViTs), which have proven effective in capturing long-range dependencies in sequential data. The predictor is a narrow transformer network that maps the representations of visible patches to the representations of masked patches~\cite{bardes2024revisiting}.

We process the EEG sequence by first dividing frames into non-overlapping spatiotemporal patches, also referred to as volumes or tubelets. Each patch captures a small region of the EEG signal across several channels and time points. We then apply a patch embedding layer, combining convolutional operations with linear projections, to map each patch to a fixed-dimensional embedding vector.

The X-encoder processes the sequence of patch embeddings after masking a portion of the patches. The masking strategy is crucial for creating the prediction task. We adapt V-JEPA's multi-block masking approach, which involves masking large, spatially contiguous blocks that span the entire temporal dimension of the sequence. This strategy encourages the model to learn long-range spatiotemporal dependencies. During masking, the aspect ratio for the sampled blocks is randomly chosen in a certain range (see Table~\ref{hyper_params}).

The predictor network takes the output of the X-encoder and a set of learnable mask tokens as input. The mask tokens contain positional embeddings that indicate the location of the masked patches in the sequence. The predictor outputs a representation for each masked patch, aiming to predict the corresponding representation from the Y-encoder.

The Y-encoder processes the entire unmasked sequence of patch embeddings, providing the target representations for the predictor. To prevent representation collapse, we use an exponential moving average (EMA) of the X-encoder weights to update the Y-encoder. This approach ensures that the target representations evolve more slowly than the predicted representations, encouraging the model to learn meaningful features.

%\vspace{.5cm}

\noindent\textbf{Pre-training Objective:}
%\hspace{0.25cm}
We train the model using a loss function similar to the one proposed in~\cite{bardes2024revisiting}, which encourages the predictor, given the representations of the unmasked tokens from X-encoder and contextual tokens, to produce representations of masked tokens, similar to the target representations of masked tokens from the Y-encoder. The loss function is formulated in Equation (1):

\begin{equation}
\min_{\theta, \phi} \left\| P_{\phi}(E_{\theta}(x), \Delta y) - \text{sg}(\overline{E}_{\theta}(y)) \right\|_1 \tag{1}
\end{equation}

\noindent Where, $E_{\theta}$ is the visual encoder, and $P_{\phi}$ is the predictor network. The term $\Delta y$ represents the spatiotemporal positions of $y$, providing context to the predictor about the transformations between $x$ and $y$. The stop-gradient operation, marked by $\text{sg}(\cdot)$, prevents backpropagation through its argument, ensuring that the gradient does not flow back through $\overline{E}_{\theta}$. The goal is to make the prediction $P_{\phi}(E_{\theta}(x), \Delta y)$ close to the target representation $\overline{E}_{\theta}(y)$. The $L_1$ loss (mean absolute error) measures the difference between these predictions and targets.

This approach prevents trivial solutions (e.g., the encoder producing a constant representation) by using the exponential moving average (EMA) of the encoder $\overline{E}_{\theta}$ and the stop-gradient operation. This strategy ensures that the predictor is always ahead of the encoder, which helps prevent representation collapse.

%%%%%%%%%%%%%%%%%%%%%%%%%%%%%%%%%%%%%%%%%%%%%%%%%%%%%%%%%%%%%%%%%%%%%%%%
\section{Experiments}\label{exp}

\subsection{Experimental Setting}

\noindent\textbf{Datasets:}
%\hspace{0.25cm} 
We use three publicly available datasets, namely TUAB~\cite{harati2014tuh}, NMT~\cite{khan2022nmt}, and the EEG dataset from the General Hospital of Thessaloniki. 

\textit{TUH Abnormal EEG Corpus (TUAB)} corresponds to the abnormal EEG dataset, a subset of the Temple University Hospital(TUH) EEG corpus. The dataset contains 2,993 recordings from 2,329 unique subjects. The recordings are annotated as \textit{abnormal/pathological}  or \textit{normal} by experts. The subject age ranges widely from 7 days to 96 years old. The dataset includes  46.7\% male and 53.3\% female subjects. Among the EEG recordings, 50.85\% of the male recordings are annotated as abnormal, while 44.75\% of the female recordings are annotated as abnormal.

\textit{NMT} dataset contains 2417 EEG recordings from a South Asian population, also divided into normal and abnormal/pathological classes. The dataset includes subjects with ages ranging from under 1 year to 90 years old. Of the records, male subjects stand at 66.56\% and females at 33.44\%. Among the male EEG recordings, 16.17\% belong to the abnormal or pathological class, while 19.18\% of the female recordings belong to this class.

\textit{General Hospital of Thessaloniki EEG dataset} contains resting-state EEG recordings collected at the General Hospital of Thessaloniki~\cite{ntetska2025complementary}. Each recording is labeled according to clinical diagnosis, including Alzheimer’s disease (AD), frontotemporal dementia (FTD), and cognitively normal controls (CN). EEG data were acquired using a 19-channel cap, arranged according to the international 10–20 electrode placement system, with a referential montage referenced to Cz. All recordings were sampled at 500 Hz under resting-state, eyes-open conditions. The dataset comprises recordings from 36 individuals with AD, 23 with FTD, and 29 healthy controls
%\vspace{.5cm}

\noindent\textbf{Preprocessing:}
%\hspace{0.25cm}
We apply some basic preprocessing steps on the EEG signals to prepare them as input for our model. We crop the recordings to a fixed 5-minute time length and select a common subset of 19 channels. We downsampled the signals to 100Hz. We then apply bandpass filtering between 1-40 Hz. Afterward, we use a sliding window technique to produce 5-second overlapping windows of size $(19 \times 500)$,  transforming the EEG signal into a 3D tensor of size $118 \times 19 \times 500$. As the last step, we normalize the EEG data channel-wise to achieve a zero mean and unit standard deviation.

%\vspace{.5cm}

\noindent\textbf{Network Architecture:}
%\hspace{0.25cm}
We employ the vision transformer (ViT) model ~\cite{dosovitskiy2020image} as the backbone network architecture. It takes in a 1D sequence of tokens with $d$ dimension and outputs a vector of $d$ dimensions for each token. The encoder is a standard ViT network with $L$ number of transformer layers. The predictor is a narrow ViT with an embedding dimension of 384 that gives an embedding vector for each mask token; see Figure~\ref{pipeline}. For simplicity, during our experiments, we keep the predictor's number of heads (NH) the same as that of the encoders´. 

Following the literature, we use the following naming scheme for the ViT models: ViT-Base (ViT-B, $L=12$, $NH=12$, $d=768$) with 85 M parameters, ViT-Medium (ViT-M, $L=12$, $NH=6$, $d=384$) with 21 M parameters, and ViT-Small (ViT-S, $L=12$, $NH=3$, $d=192$)  with 5 M parameters, where $L$ is the number of transformer layers, each with a self-attention block of $NH$ heads. To indicate the tubelet size while patching, we extend the naming scheme, for example, ViT-B/$4 \times 30 \times 2$ denotes a ViT-Base backbone with the tubelet size of ($h \times w \times t = 4 \times 30 \times 2$). We tested various ViT encoder sizes (base, medium, small) to evaluate the impact of model capacity on representation learning; refer to  Table~\ref{hyper_params} for more details. In all our experiments, the tubelet height ($h$) and width($w$) differ due to the shape of our EEG input signals´ height and width ($19 \times 500$). We also probe with the time dimension $t$ while patching. 

%\vspace{.5cm}

\noindent\textbf{Pre-training:} We combine two publicly available datasets, namely TUH~\cite{harati2014tuh} and NMT~\cite{khan2022nmt}, to create an unsupervised EEG pre-training dataset that we refer to as EEGComb2. The resulting EEGComb2 dataset for self-supervised pertaining contains EEG recordings from 4438 subjects. We pretrain EEG-VJEPA model with the vision transformer (ViT) as the encoder´s backbone on EEGComb2 dataset. We train a ViT-B/$h\times w \times t$, a ViT-M/$h\times w \times t$, and a ViT-S/$h\times w \times t$ transformer models with various tublet sizes. We use different batch sizes (4,5,6,10,30,40) during experiments; see the combination of different hyperparameters during pre-training in detail in Table~\ref{hyper_params}. Key hyperparameters include:

\noindent\textbf{Spatial Augmentation}: Techniques such as resizing and noise addition to enhance robustness.

\noindent\textbf{Frame Flipping}: Random temporal flipping to reduce sensitivity to signal direction.

\noindent\textbf{Patch Size and Tubelet Size}: Variations in spatial and spatiotemporal patch sizes to balance local and global feature extraction.

\noindent\textbf{Number of Frames and Clips}: Adjustments in sequence length and segment coverage for capturing temporal dependencies.

\noindent\textbf{Random Resizing}: Random aspect ratio and scaling for 3D masking.

\noindent\textbf{Encoder Size}: ViT-S (5M params), ViT-M (21M params), and ViT-B (85M params) to evaluate the impact of model capacity.

The pre-training details, including predictor depth, sampling rates, and augmentation strategies, are outlined in Table~\ref{tab:hyperparameters}. These configurations highlight the versatility and scalability of the EEG-VJEPA framework for self-supervised representation learning.

\begin{table*}[ht]
     \caption{Summary of the main hyper-parameters tuned for different model versions during pre-training.}
    \centering
    \arrayrulecolor{headerbg}
    \resizebox{\textwidth}{!}{%
    \begin{tabular}{|c|c|c|c|c|c|c|c|c|c|c|}
        \hline
        \rowcolor{headerbg}
        \textcolor{White}{\textbf{Version}} & \textcolor{White}{\textbf{Augment}} & \textcolor{White}{\textbf{Batch}} & \textcolor{White}{\textbf{Encoder size}} & \textcolor{White}{\textbf{Predictor depth}} & \textcolor{White}{\textbf{Sampling}} & \textcolor{White}{\textbf{Frames}} & \textcolor{White}{\textbf{Clips}} & \textcolor{White}{\textbf{Tubelet Temp. size}} & \textcolor{White}{\textbf{rrar/rrs}} & \textcolor{White}{\textbf{Patch size}} \\
        \hline
        1 & flip, spatial & 4 & small & 4 & 2 & 16 & 3 & 2 & (0.75, 1.35)/(0.3, 1) & (2, 30) \\
        \hline
        2 & spatial & 4 & small & 4 & 2 & 16 & 3 & 2 & (0.75, 1.35)/(0.3, 1) & (2, 30) \\
        \hline
        3 & - & 40 & base & 12 & 2 & 16 & 3 & 2 & - & (2, 30) \\
        \hline
        4 & spatial & 10 & base & 12 & 2 & 16 & 3 & 2 & (0.75, 1.35)/(0.3, 1) & (2, 30) \\
        \hline
        4.1 & spatial & 10 & base & 12 & 2 & 16 & 3 & 2 & (0.75, 1.35)/(0.3, 1) & (2, 50) \\
        \hline
        4.2 & spatial & 10 & base & 12 & 2 & 16 & 3 & 2 & (0.75, 1.35)/(0.3, 1) & (2, 15) \\
        \hline
        4.3 & spatial & 10 & base & 12 & 2 & 16 & 3 & 2 & (0.75, 1.35)/(0.3, 1) & (4, 30) \\
        \hline
        5 & flip, spatial & 10 & base & 12 & 2 & 16 & 3 & 2 & (0.75, 1.35)/(0.3, 1) & (2, 30) \\
        \hline
        6 & spatial, noise, scale & 10 & base & 12 & 2 & 16 & 3 & 2 & (0.75, 1.35)/(0.3, 1) & (2, 30) \\
        \hline
        7 & spatial & 10 & base & 12 & 1 & 64 & 1 & 8 & (0.75, 1)/(0.3, 1) & (2, 30) \\
        \hline
        8 & spatial & 10 & base & 12 & 3 & 32 & 1 & 4 & (0.75, 1.35)/(0.3, 1) & (2, 30) \\
        \hline
        8.1 & spatial & 10 & base & 12 & 3 & 32 & 1 & 4 & (0.75, 1.35)/(0.3, 1) & (4, 30) \\
        \hline
        8.2 & spatial & 30 & base & 12 & 3 & 32 & 1 & 4 & (0.75, 1.35)/(0.3, 1) & (4, 30) \\
        \hline
        9 & spatial & 6 & medium & 6 & 2 & 16 & 3 & 2 & (0.75, 1.35)/(0.3, 1) & (2, 30) \\
        \hline
        9.1 & spatial & 30 & medium & 6 & 3 & 32 & 1 & 4 & (0.75, 1.35)/(0.3, 1) & (4, 30) \\
        \hline
        9.2 & spatial & 10 & medium & 6 & 3 & 32 & 1 & 4 & (0.75, 1.35)/(0.3, 1) & (4, 30) \\
        \hline
        9.3 & spatial & 10 & medium & 6 & 3 & 32 & 1 & 4 & (0.75, 1)/(0.75, 1) & (4, 30) \\
        \hline
        10 & spatial & 5 & small & 4 & 1 & 16 & 6 & 2 & (0.75, 1)/(0.3, 1) & (2, 30) \\
        \hline
        10.1 & spatial & 10 & small & 4 & 3 & 32 & 1 & 4 & (0.75, 1)/(0.3, 1) & (4, 30) \\
        \hline
        10.2 & spatial & 30 & small & 4 & 3 & 32 & 1 & 4 & (0.75, 1)/(0.3, 1) & (4, 30) \\
        \hline
        11 & spatial & 10 & base & 12 & 1 & 16 & 6 & 2 & (0.75, 1.35)/(0.3, 1) & (2, 30) \\
        \hline
    \end{tabular}
    }
    \vspace{1em}
   
    \label{tab:hyperparameters}
\end{table*}

\begin{figure*}[ht]
    \centering
    \includegraphics[width=\textwidth]{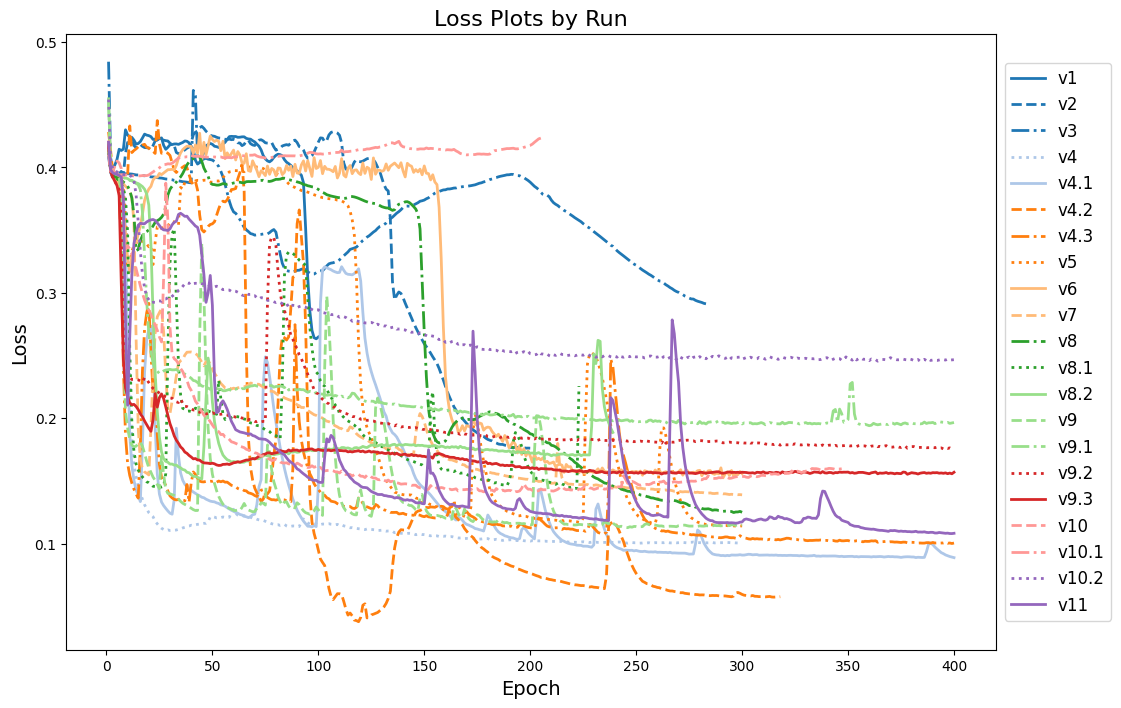}
    %\vspace{-1em}
    \caption{The pre-training loss ofThe LaBraM~\cite{jiang2024large} model is trained over 20 EEG datasets using vector quantized neural spectrum prediction to generate neural vocabularyd from Table~\ref{tab:hyperparameters}. }
    \label{loss_runs}
     \vspace{2em}
\end{figure*}

%\vspace{.5cm}
\noindent\textbf{Baselines:} 
%\hspace{0.20cm} 
Our self-supervised baseline consists of EEG2REP \cite{foumani2024eeg2rep}, LaBraM~\cite{jiang2024large}, and the contrastive learning (CL) based model \cite{hojjati2023multi}, both trained on the TUAB dataset~\cite{obeid2016temple}. The EEG2REP model is based on the I-JEPA~\cite{assran2023self} model, while the contrastive learning model is based on the concept of multi-view learning~\cite{kumar2022muleeg} with ViT as its backbone. The LaBraM~\cite{jiang2024large} is also a transformer based model pre-trained over 20 EEG datasets by vector quantized neural spectrum prediction, where the tokenizer predicts the Fourier spectrum of the original signal. Additionally, we benchmark against supervised methods, including BSVT~\cite{khadka2024inducing}, a vision transformer-based approach, and Chrononet~\cite{roy2019chrononet}, an RNN-based model. We also include a baseline SVM classifier~\cite{rostamikia2024eeg}, trained on a combination of time-domain features, frequency-domain features, connectivity metrics, and complexity-based measures, in our evaluation on the smaller, independent dataset from the General Hospital of Thessaloniki.
%\vspace{.5cm}

\noindent\textbf{Implementaiton Details:}
%\hspace{0.25cm}
We used the AdamW optimizer \cite{loshchilov2017decoupled} with betas parameters set to (0.9, 0.999) and an epsilon value of 1e-8. The learning rate followed a WarmupCosineSchedule, which included an initial warmup phase (about 40 epochs) to a starting learning rate (0.0002) followed by a cosine decay to the reference learning rate (0.000625) and towards the final learning rate (1e-6) during training. The weight decay was managed using a CosineWDSchedule (0.04, 0.4). A momentum scheduler was implemented with values ranging from 0.998 to 1.0 for the exponential moving average (EMA) of the encoder's weights, ensuring that the target encoder's weights evolve more slowly and stably than the source encoder's weights. Gradient clipping with a maximum norm of 10.0 was applied to prevent exploding gradients. We ran the pre-training runs on 5 Nvidia V100-32GB GPUs for up-to 400 epochs, and the fine-tuning and evaluation runs on a single V100-32GB GPU for up to 500 epochs.

\begin{figure*}[t]
    \centering
    \includegraphics[width=\textwidth,height=4cm]{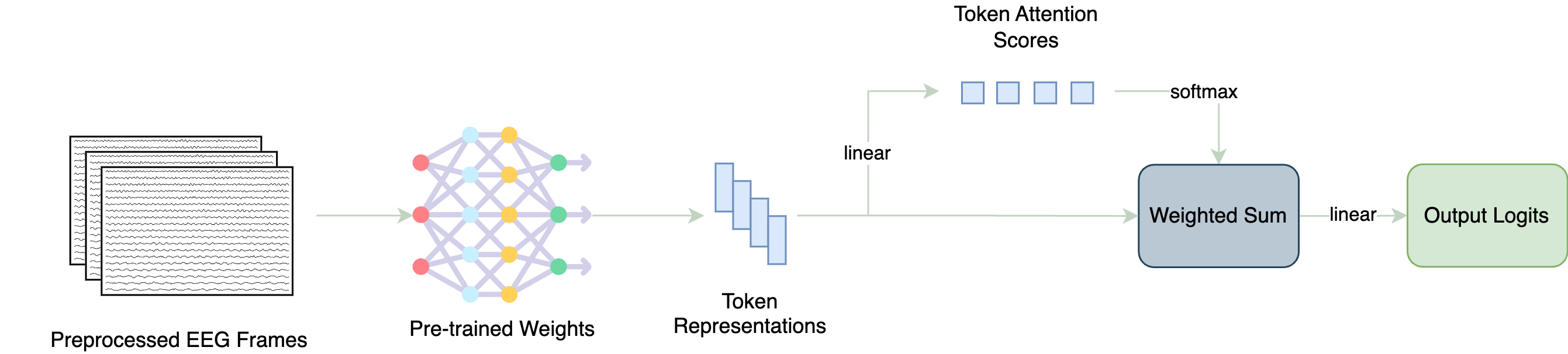}
    %\vspace{-1em}
    \caption{Inference. The pre-processed EEG signals are input into the pre-trained EEG-VJEPA model. The query token representations from the pre-trained encoder pass through a cross-attention layer, with its output added to the query tokens via a residual connection. Finally, this combined output is fed into a linear classifier.}
    %\vspace{-1em}
    \label{eval_pipeline}
\end{figure*}

\begin{figure*}[htbp]
    \centering
    \begin{subfigure}{0.32\textwidth}
        \centering
        \includegraphics[trim={20mm 10mm 20mm 10mm}, clip, width=\linewidth]{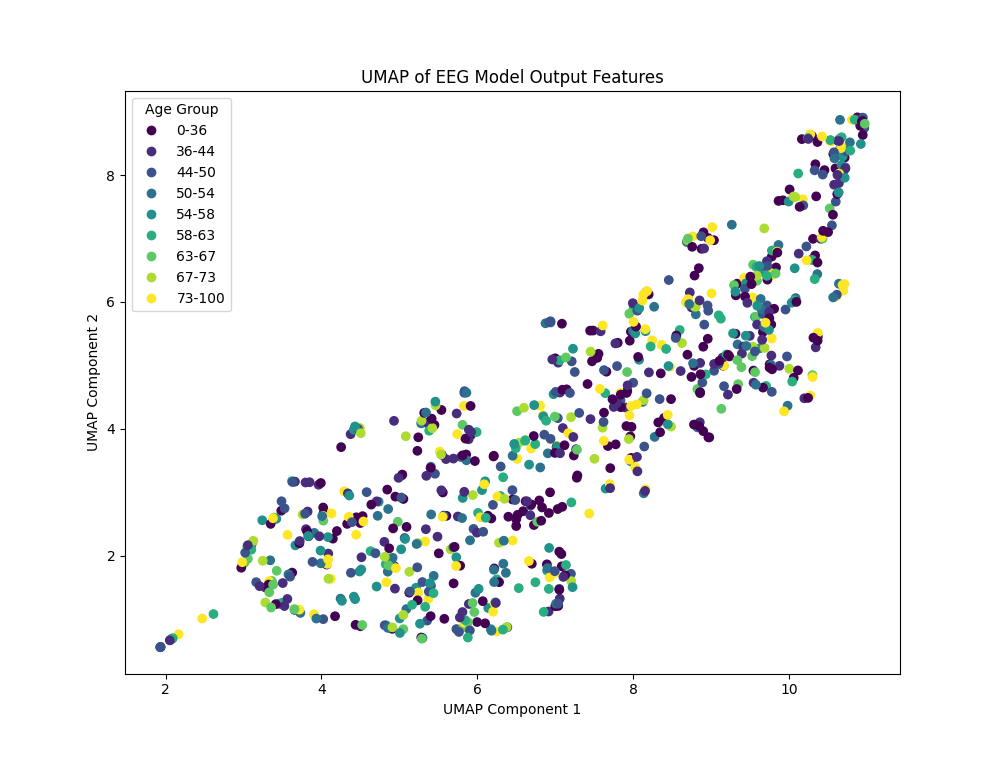}
        \caption{}
        \label{fig:subfig1}
    \end{subfigure}
    %\hspace{-2em} 
    \begin{subfigure}{0.32\textwidth}
        \centering
        \includegraphics[trim={20mm 10mm 20mm 10mm}, clip, width=\linewidth]{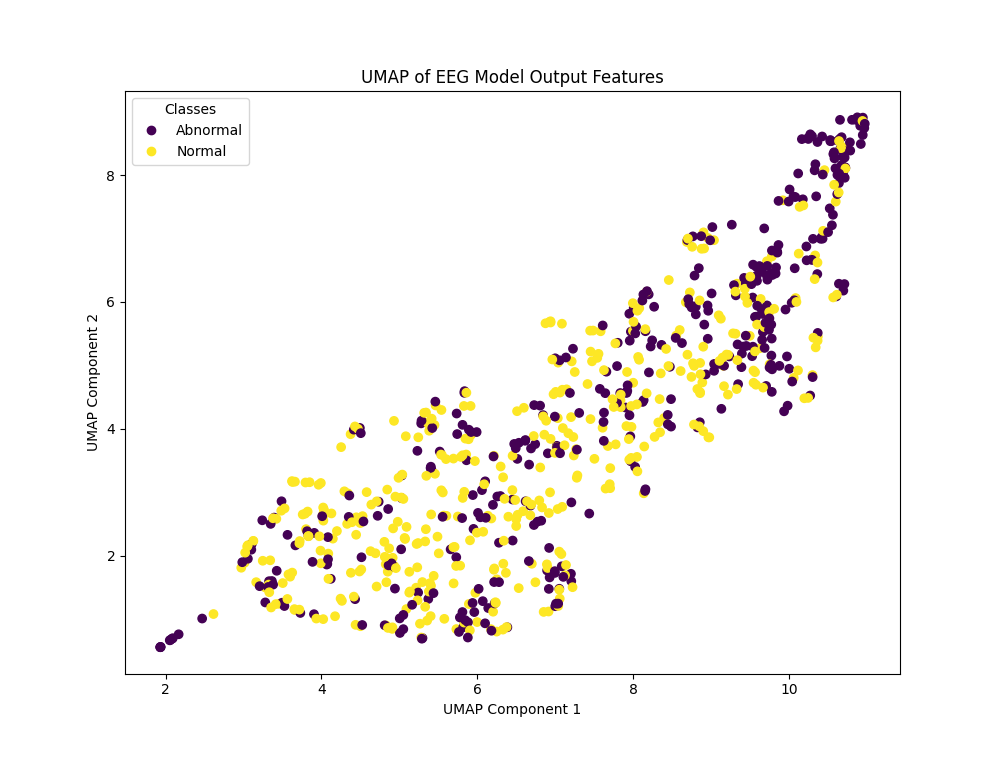}
        \caption{}
        \label{fig:subfig2}
    \end{subfigure}
    %\hspace{-1em} % Optional horizontal space between subfigures
    \begin{subfigure}{0.32\textwidth}
        \centering
        \includegraphics[trim={20mm 10mm 20mm 10mm}, clip, width=\linewidth]{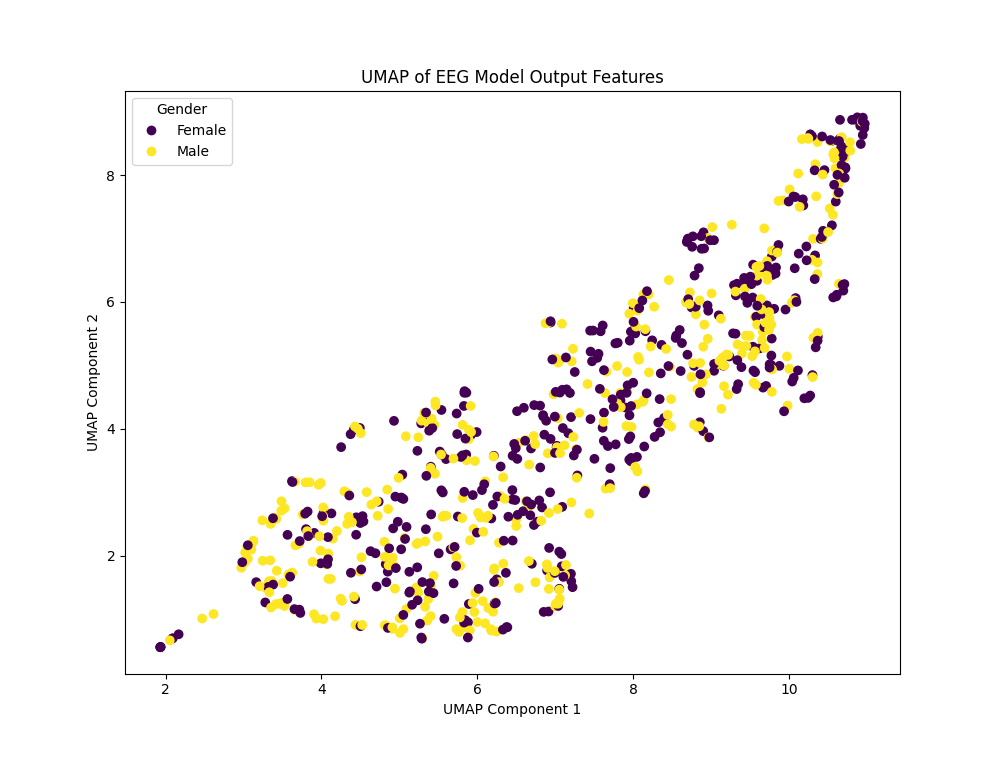}
        \caption{}
        \label{fig:subfig3}
    \end{subfigure}
    \vspace{1em}
    \caption{UMAP-based 2D visualization of feature embeddings of EEG-VJEPA. (a) Age-related clusters appear in the embeddings.   (b) A global structure emerges in the embeddings, grouped by pathological labels (Abnormal/Normal). (c) The gender-related global structure in the embeddings. }
    \vspace{2em}

    \label{fig:fig_structure}
\end{figure*}

\begin{figure*}[htbp!]
    \centering
    \includegraphics[trim={3mm 0mm 0cm 3mm}, clip, width=0.85\textwidth, height=0.85\textheight]{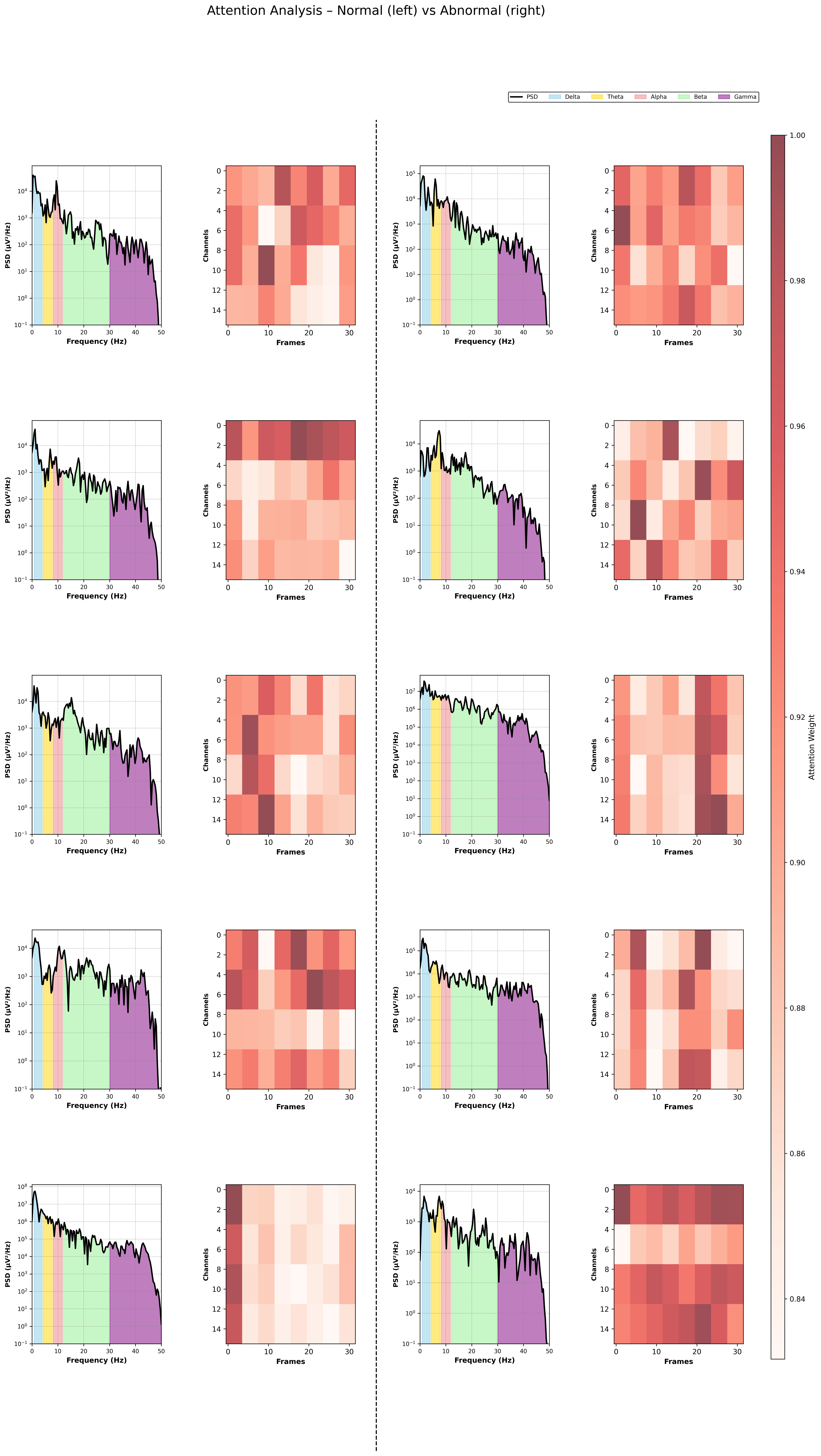}
    %\vspace{-1em}
    %\vspace{-1em}
    \caption{EEG-VJEPA locates regions of interest using spatiotemporal tokens. We roll out attention weights through all layers to visualize the attention along the EEG channel and time dimensions in 2D. The corresponding PSD plots show how EEG signal power is distributed across frequency bands, highlighting differences between normal (left side)  and abnormal (right side) samples, across five different subjects in each class.}
    
    \label{attn_rollout}
    \vspace{1em}
    
\end{figure*}

\subsection{Evaluation on TUAB dataset}\label{eval}

We evaluate the models on a subset of the TUAB dataset~\cite{obeid2016temple}, which includes 276 normal and 270 abnormal recordings for training, and 150 normal and 126 abnormal recordings for validation. We implement an attention-based linear evaluation protocol to assess the quality of our pre-trained EEG representations. After freezing the weights of the pre-trained encoder, we add the classification layer on top of the network. As shown in Figure~\ref{eval_pipeline}, during the forward pass, the layer calculates the attention weights for each token in the input sequence, applies these weights to create a weighted sum of the embeddings, and then passes this sum through a final linear layer for classification. This approach allows the classifier to focus on different parts of the input sequence, each with varying importance. It provides a better assessment of the model's performance compared to global averaging and a linear classification head. 

\begin{table}[t]
       \caption{Comparison of EEG-VJEPA models with the state-of-the-art baselines; including a self-supervised contrastive learning (CL) method evaluated with accuracy and F1-score, and Chrononet evaluated with accuracy. Performance results are presented as mean and standard deviation computed over five independent runs. The EEG-VJEPA variant employing ViT-M/$4 \times 30 \times 4$ achieves performance comparable to the fully supervised Chrononet model and significantly outperforms the contrastive learning baseline. Additionally, the EEG-VJEPA model with ViT-B/$4 \times 30 \times 2$ demonstrates competitive results, closely matching the baseline model. 
    }
    \centering
    \arrayrulecolor{headerbg} 

    \begin{tabular}{|c|c|c|}
        \hline
        \rowcolor{headerbg} 
        \textcolor{White}{\textbf{Method}} & \textcolor{White}{\textbf{Accuracy}} & \textcolor{White}{\textbf{F1}}  \\
        \hline
         CL Model~\cite{hojjati2023multi}\(^*\) & 80.55\% & 79.3\% \\
        \hline
        CL Model~\cite{hojjati2023multi}\(^\dagger\)  & 81.9\% & 81.5\% \\
        \hline
        BSVT~\cite{khadka2024inducing}\  & 82.67 \% & 77.32 \%\\
        \hline
        Chrononet~\cite{roy2019chrononet}\  & 86.57 \% & - \\
        \hline
        
       \rowcolor{lightblue} ViT-B/$4 \times 30 \times 2$  \(^*\) & 81.20\% $\pm{0.3}$& 81.0\%$\pm{0.2}$ \\
        \hline
        \rowcolor{lightblue}ViT-M/$4 \times 30 \times 4$  \(^*\) & 83.30\% $\pm{0.3}$  & 82.4\%$\pm0.2$ \\
        \hline
       \rowcolor{lightblue} ViT-B/$4 \times 30 \times 2$ \(^\dagger\)  & 80.55\%$\pm{0.4}$ & 80.1\%$\pm{0.3}$ \\
        \hline
        \rowcolor{lightblue}ViT-M/$4 \times 30 \times 4$  \(^\dagger\)  & \textbf{85.80}\%$\pm{0.5}$ & \textbf{85.60\%}$\pm{0.4}$\\
        \hline
       
    \end{tabular}
    \vspace{1em}

    \caption*{\(^*\)Frozen evaluation, \(^\dagger\)Fine-tuning.}
    \label{tab:comp_CL}

\end{table}

\begin{table}[t]
     \caption{Comparison of EEG-VJEPA models with the state-of-the-art baseline models trained with JEPA for accuracy and AUROC.Performance results are presented as mean and standard deviation computed over five independent runs. The EEG-VJEPA, based on the ViT-M/$4 \times 30 \times 4$ network, outperforms the baseline models in both fine-tuning and frozen evaluation setups. Additionally, EEG-VJEPA with ViT-B/$4 \times 30 \times 2$ also demonstrates competitive performance, matching or even surpassing the baseline model.}
    \centering
    \begin{tabular}{|c|c|c|}
        \hline
        \rowcolor{headerbg} % Apply predefined color to header
        \textcolor{White}{\textbf{Method}} & \textcolor{White}{\textbf{Accuracy}} & \textcolor{White}{\textbf{AUROC}} \\
        \hline
        EEG2Rep~\cite{foumani2024eeg2rep} \(^*\) & 76.6\% & 83.2\% \\
        \hline
        EEG2Rep~\cite{foumani2024eeg2rep} \(^\dagger\)  & 80.5\% & 88.4\% \\
        \hline
        LaBraM~\cite{jiang2024large} \(^\dagger\)  & 82.58\% & 92.04\% \\
        \rowcolor{lightblue} ViT-B/$4 \times 30 \times 2$  \(^*\) & 81.20\%$\pm{0.3}$ & 87.9\%$\pm{0.1}$ \\
        \hline
        \rowcolor{lightblue} ViT-M/$4 \times 30 \times 4$  \(^*\) & {83.30\%}$\pm{0.3}$ & 87.7\%$\pm{0.2}$ \\
        \hline
        \rowcolor{lightblue} ViT-B/$4 \times 30 \times 2$ \(^\dagger\) & 80.55\%$\pm{0.4}$ & 86.8\%$\pm{0.2}$\\
        \hline
        \rowcolor{lightblue} ViT-M/$4 \times 30 \times 4$  \(^\dagger\) & \textbf{85.80\%}$\pm{0.5}$ & \textbf{88.5\%}$\pm{0.2}$ \\
      
        \hline
    \end{tabular}
    
    \vspace{1em}

    \caption*{\(^*\)Frozen evaluation, \(^\dagger\)Fine-tuning.}
    \label{tab:comp_JEPA}

    %\vspace{-2em}
\end{table}
%\vspace{-1em}

\noindent\textbf{Frozen Evaluation:}
We utilize the frozen evaluation pipeline illustrated in Figure~\ref{eval_pipeline} with a subset of the TUAB dataset. We compare EEG-VJEPA with the baseline models, LaBraM~\cite{jiang2024large}, EEG2Rep~\cite{foumani2024eeg2rep}, BSVT~\cite{khadka2024inducing}, Chrononet~\cite{roy2019chrononet}, and the CL model~\cite{hojjati2023multi}. The EEG-VJEPA with ViT-M/$4 \times 30 \times 4$  backbone, under frozen evaluation setup, achieves an accuracy of 83.30\%$\pm{0.3}$, an F1 score of 82.4\%$\pm{0.2}$, and an AUROC of 87.7\%$\pm{0.2}$ over five independent runs. We can observe the increase in performance gap in comparison with the baseline models, particularly with the CL model, BSVT, LaBraM, and EEG2Rep (see Table~\ref{tab:comp_CL} and Table~\ref{tab:comp_JEPA} ).

%\vspace{.5cm}

\noindent\textbf{Fine-Tuning:}
We evaluate the EEG-VJEPA model by fine-tuning on a subset of the TUAB dataset as mentioned in Section~\ref{eval}. The medium version of the EEG-VJEPA model with ViT-M/$4 \times 30 \times 4$  achieves an accuracy of 85.80\%$\pm{0.5}$, F1 score of 85.6\%$\pm{0.4}$, and an AUROC of 88.5\%$\pm{0.2}$ under fine-tuning settings. It outperforms the state-of-the-art baseline models and is on par with  Chrononet, a model trained under a fully supervised setting. We also include in the comparison the EEG-VJEPA model with ViT-B/$4 \times 30 \times 2$ backbone (see Table~\ref{tab:comp_CL} and Table~\ref{tab:comp_JEPA} ).

% In the fine-tuning setting, there is not much difference in performance between EEG-VJEPA and the baseline model.

\noindent\textbf{EEG-VJEPA learns meaningful representations: }To observe different structures learned by EEG-VJEPA, we uncover the learned embeddings and map their relationship with pathological and demographic attributes such as gender (male/female), age groups. The 384-dimensional embeddings obtained on the TUAB dataset are projected onto a two-dimensional space with Uniform Manifold Approximation~\cite{mcinnes2020umapuniformmanifoldapproximation}. 
 In Figure~\ref{fig:subfig1}, we observe age-associated local structures with a gradient. The younger age groups tend to cluster towards the left region of the plot, and the older age groups lean more towards the right, although with some overlap. Similarly, Figure~\ref{fig:subfig2} reveals a global structure based on pathology. While observing Figure~\ref{fig:subfig1} and Figure~\ref{fig:subfig2} simultaneously, we can observe an alignment where normal EEG samples more often corresponded with younger age groups region and abnormal samples with the older groups region in the plot. The gender-related global structure in Figure~\ref{fig:subfig3} shows a distribution with some overlap between male and female groups. This suggests that the model captured gender-specific features, however, there are shared physiological characteristics in the EEG signals across genders. It is worth noting that factors such as dimensionality reduction, generalized features, and the presence of noise may also contribute to the variability and overlap in the observed clusters. 

To compute the attention maps and visualize the attention scores across the whole network, from the spatiotemporal input patches to final representations, we utilized the well-known method called Attention Rollout introduced in~\cite{Abnar2020QuantifyingAF} to compute the attention flow across layers, which has been previously used to visualize attention maps in vision transformers~\cite{dosovitskiy2021AnImage}. This can be achieved by a recursive multiplication of the attention matrices, layer by layer, while accounting for residual connections. Equation (2) shows the attention rollout process. 

\begin{align}
\hat{A}^{(l)} &= A^{(l)} + I, \nonumber \\
\widetilde{A}^{(l)}_{ij} &= \frac{\hat{A}^{(l)}_{ij}}{\sum_{k} \hat{A}^{(l)}_{ik}}, \nonumber \\
A^{\mathrm{rollout}} &= \prod_{l=1}^{L} \widetilde{A}^{(l)}.
\tag{2}
\end{align}

Where, $\mathbf{A}^{\mathrm{rollout}}$ is the final attention rollout matrix, $\mathbf{A}^{(l)}$ is the raw attention matrix at layer $l$, $\mathbf{I}$ is the identity matrix (residual connection) and $L$ is the total number of layers.

%\begin{itemize}
  %\item $\mathbf{A}^{\mathrm{rollout}}$ is the final attention rollout matrix.
  %\item $\mathbf{A}^{(l)}$ is the raw attention matrix at layer $l$.
  %\item $\mathbf{I}$ is the identity matrix (residual connection).
  %\item $L$ is the total number of layers.
%\end{itemize}

As shown in Figure~\ref{attn_rollout}, the 2D attention maps demonstrate the model's ability to focus on spatiotemporal regions of interest within the input sequences. By propagating the attention through all layers, we visualize the dynamic focus of the model on diagnostically relevant spatial channels and temporal frames. The power spectral density (PSD) plots show how EEG signal power is distributed across different frequency bands, highlighting spectral differences between normal and abnormal samples that can help in distinguishing brain states. We computed each band’s relative power from the PSD, guided by the model’s attention maps, for both normal and abnormal recordings. In the normal condition, beta rhythms accounted for 27.81\% of total power; in the abnormal condition, beta power fell to 11.48\%. This decrease in faster oscillations echoes prior reports of diminished beta activity in pathological EEGs~\cite{liu2021eeg}.

In the context of EEG applications, this analysis highlights that the model effectively captures physiologically and clinically relevant features embedded in the data, which are crucial in the development of interpretable biomarkers. 
 
All visualizations are made using the ViT-M/$4 \times 30 \times 4$ model.

%%%%%%%%%%%%%%%%%%%%%%%%%%%%%%%%%%%%%%%%%%%%%%%%%%%%%%%%%%%%%%%%%%%%%%%%%

\subsection{Evaluation on the General Hospital of Thessaloniki EEG Dataset}\label{GHTeval}
To assess the generalization capabilities of EEG-VJEPA on a smaller, independent dataset, we utilized the General Hospital of Thessaloniki EEG dataset (88 subjects), comprising individuals diagnosed with Alzheimer’s disease (AD), frontotemporal dementia (FTD), and cognitively normal controls (CN). For binary classification, we merged the AD and FTD groups into a single dementia category.

Following the evaluation method shown in Figure~\ref{eval_pipeline}, we fine-tune EEG-VJEPA on the General Hospital of Thessaloniki EEG dataset. The EEG-VJEPA model based on ViT-M/$4 \times 30 \times 4$  achieves an accuracy of 83.34\%$\pm{0.2}$, F1 score of 83.48\%$\pm{0.2}$ under fine-tuning settings. While the performance is strong, it remains slightly below the baseline SVM model (see Table~\ref{tab:comp_alz88}). The performance gap likely reflects the small dataset size and the handcrafted features tailored to dementia in the SVM approach. In contrast, our method requires no manual feature design and learns general EEG patterns through pretraining, offering a flexible foundation that may improve with larger or more diverse pretraining data.

\begin{table}[htb!]
     \caption{Classification performance of EEG-VJEPA on the General Hospital of Thessaloniki EEG dataset (88 subjects). Results are reported for the ViT-M/$4 \times 30 \times 4$ configuration under fine-tuning and frozen settings.}
    \centering
    \begin{tabular}{|c|c|c|c|}
        \hline
        \rowcolor{headerbg} % Apply predefined color to header
        \textcolor{White}{\textbf{Method}} & \textcolor{White}{\textbf{Accuracy}} & \textcolor{White}{\textbf{Recall}} &\textcolor{White}{\textbf{F1}} \\
        \hline
        SVM~\cite{rostamikia2024eeg} \(^*\) & 93.5\% & 90.0\% & -        
        \\
        \hline
        \rowcolor{lightblue} ViT-M/$4 \times 30 \times 4$  \(^*\) & {67.00\%}$\pm{0.2}$ & 54.66\%$\pm{0.2}$& 53.21\%$\pm{0.2}$ \\
        
        \hline
        \rowcolor{lightblue} ViT-M/$4 \times 30 \times 4$  \(^\dagger\) & \textbf{83.34\%}$\pm{0.2}$ & \textbf{83.70\%}$\pm{0.2}$ &\textbf{83.48\%}$\pm{0.2}$ \%\\
      
        \hline
    \end{tabular}
    \vspace{1em}

    \caption*{\(^*\)Frozen evaluation, \(^\dagger\)Fine-tuning.}
    \label{tab:comp_alz88}

\end{table}

\section{Discussion}

EEG signals exhibit unique spatial and temporal characteristics that provide critical insights into dynamic brain activity, essential for effective EEG analysis and BCI applications. To address these challenges, we proposed EEG-VJEPA, a self-supervised model designed to learn robust representations of EEG data. Pre-trained on two diverse datasets, the proposed model consistently outperforms baseline methods, notably surpassing the state-of-the-art LaBraM~\cite{jiang2024large}, which was trained on 20 different EEG datasets, and achieving performance on par with the fully supervised ChronoNet model~\cite{roy2019chrononet}. It also exhibits a substantial performance advantage over self-supervised approaches such as contrastive learning~\cite{hojjati2023multi} and the JEPA-based EEG2REP model~\cite{foumani2024eeg2rep}, underscoring the benefit of video-inspired joint embedding predictive modeling for EEG signals. We further evaluated EEG-VJEPA on an independent, smaller clinical dataset from the General Hospital of Thessaloniki, focusing on the binary classification of dementia versus healthy controls. Despite the limited sample size, EEG-VJEPA demonstrated strong generalization capabilities and competitive performance compared to the supervised baseline, highlighting its potential applicability in real-world clinical scenarios where labeled data are scarce.

Several factors significantly influenced the model's performance. Patch sizes, which control the granularity of spatiotemporal embeddings, were crucial; configurations like (4, 4, 30) and (2, 4, 30) balanced stability and generalization, whereas smaller patches often led to overfitting. Spatial augmentations such as noise addition and scaling further enhanced generalization, while batch size adjustments stabilized training by improving gradient convergence, particularly in smaller models. Larger encoders, such as ViT-M and ViT-B, captured more complex patterns in EEG data, but the benefits diminished with overly large architectures, likely due to dataset size limitations.

Interestingly, the lowest pre-training loss did not always translate to the best validation performance, indicating that excessive optimization on the pretext task could lead to overfitting and hinder generalization. Fine-tuning strategies also played a pivotal role: for smaller datasets, restricting fine-tuning to the last layer reduced overfitting, while full fine-tuning with larger datasets benefited from lower learning rates and partial unfreezing.

Overall, EEG-VJEPA effectively captured meaningful patterns from unlabeled EEG data, as evidenced by consistent reductions in pre-training loss and superior frozen evaluation results. These findings highlight the model’s ability to generalize well to unseen tasks, demonstrating its potential for advancing EEG representation learning.

\section{Conclusion}

In this work, we introduced \textbf{EEG-VJEPA}, a novel self-supervised learning framework for EEG representation learning, which adapts the Video Joint Embedding Predictive Architecture (V-JEPA) to model multichannel EEG as spatiotemporal sequences. By leveraging masked latent prediction and a Vision Transformer backbone, EEG-VJEPA learns semantically meaningful and structured representations without the need for expert-labeled data. We pre-trained the model on two large-scale public EEG datasets and evaluated it on the Temple University Hospital Abnormal EEG Corpus (TUAB), a clinically realistic and heterogeneous benchmark. EEG-VJEPA outperforms prior self-supervised approaches and performs comparably to fully supervised models such as Chronet. Moreover, it demonstrates strong generalization to a smaller, independent dataset from the General Hospital of Thessaloniki. Importantly, the model's attention patterns and latent embeddings align with physiologically and clinically meaningful EEG features, offering a pathway to interpretability and trust.

Our findings position EEG-VJEPA as a strong foundation model candidate for scalable, label-efficient EEG-based applications. The architecture’s flexibility and generalization across datasets suggest its potential for diverse downstream clinical tasks such as anomaly detection, triage, and risk stratification. Furthermore, its interpretability opens avenues for integration into human-AI collaborative systems, supporting transparency and clinical decision-making.

\noindent\textbf{Limitations and Future Directions.}  
While EEG-VJEPA demonstrates strong performance, several limitations remain. First, despite pretraining on relatively large datasets, the diversity of EEG patterns encountered in real-world clinical environments remains underrepresented. Future work should systematically address cross-institutional and cross-device generalization, fairness across demographic subgroups, and robustness to both distributional shifts and adversarial perturbations. Methodologically, there is considerable scope to refine EEG-specific masking strategies, incorporate multiresolution temporal modeling, and explore hybrid architectures combining transformers with convolutional or graph-based components.

In addition, scaling the model to larger parameter regimes, akin to foundation models in vision and language domains, may yield further performance gains and uncover emergent capabilities. Given the computational demands of full fine-tuning, future efforts could also focus on parameter-efficient adaptation strategies such as adapters, low-rank adaptation (LoRA), or prompt tuning. From a deployment perspective, integrating EEG-VJEPA into real-time, resource-constrained clinical workflows will require optimizing for inference speed, memory efficiency, and energy consumption. Incorporating uncertainty quantification and systematically gathering clinician feedback will be essential for building trustworthy, human-centered decision support systems.

Finally, expanding EEG-VJEPA toward multimodal learning, combining EEG with speech, imaging, or other physiological signals, could open new avenues for holistic neurocognitive modeling and cross-modal representation learning.

\noindent\textbf{Prospect of application}. EEG-VJEPA provides a scalable and interpretable path towards a foundation model for EEG analysis, capable of learning clinically relevant features from unlabeled data. Its strong performance on a noisy, real-world dataset and alignment with neurophysiological patterns suggest applicability in clinical workflows such as EEG triage, Intensive Care Unit (ICU) monitoring, and early disease screening, such as dementia detection. With further validation and integration into clinician-facing systems, EEG-VJEPA could support human-AI collaboration in neurological diagnostics and improve access to high-quality EEG interpretation in under-resourced settings.

% \begin{ack}
% This work has received funding from the European Union's Horizon 2020 research and innovation program under grant agreement No. 964220. We conducted experiments on the Experimental Infrastructure for Exploration of Exascale Computing (eX3) system, financially supported by RCN under contract 270053.
% \end{ack}

%%%%%%%%%%%%%%%%%%%%%%%%%%%%%%%%%%%%%%%%%%%%%%%%%%%%%%%%%%%%%%%%%%%%%%%%
%\bibliographystyle{unsrtnat}%%% Use this command to include your bibliography file.
%\bibliography{ref}
\bibliographystyle{elsarticle-num}

\bibliography{ref.bib}

@article{michel2012towards,
  title={Towards the utilization of EEG as a brain imaging tool},
  author={Michel, Christoph M and Murray, Micah M},
  journal={Neuroimage},
  volume={61},
  number={2},
  pages={371--385},
  year={2012},
  publisher={Elsevier}
}

@article{mcfarland2017eeg,
  title={EEG-based brain--computer interfaces},
  author={McFarland, Dennis J and Wolpaw, Jonathan R},
  journal={current opinion in Biomedical Engineering},
  volume={4},
  pages={194--200},
  year={2017},
  publisher={Elsevier}
}

@article{jiang2024large,
  title={Large brain model for learning generic representations with tremendous EEG data in BCI},
  author={Jiang, Wei-Bang and Zhao, Li-Ming and Lu, Bao-Liang},
  journal={arXiv preprint arXiv:2405.18765},
  year={2024}
}

@article{ye2024self,
  title={Self-supervised cross-modal visual retrieval from brain activities},
  author={Ye, Zesheng and Yao, Lina and Zhang, Yu and Gustin, Sylvia},
  journal={Pattern Recognition},
  volume={145},
  pages={109915},
  year={2024},
  publisher={Elsevier}
}

@article{zhang2025psss,
  title={PSSS-EEG: A Probabilistic-masking Self-Supervised Swin-transformer model for EEG-based drowsiness recognition},
  author={Zhang, Jiaming and Zhang, Fangzuo and Wei, Hongtao},
  journal={Pattern Recognition},
  volume={158},
  pages={111005},
  year={2025},
  publisher={Elsevier}
}

@article{rostamikia2024eeg,
  title={EEG-based classification of Alzheimer’s disease and frontotemporal dementia: a comprehensive analysis of discriminative features},
  author={Rostamikia, Mehran and Sarbaz, Yashar and Makouei, Somaye},
  journal={Cognitive Neurodynamics},
  volume={18},
  number={6},
  pages={3447--3462},
  year={2024},
  publisher={Springer}
}

@article{ntetska2025complementary,
  title={A Complementary Dataset of Scalp EEG Recordings Featuring Participants with Alzheimer’s Disease, Frontotemporal Dementia, and Healthy Controls, Obtained from Photostimulation EEG},
  author={Ntetska, Aimilia and Miltiadous, Andreas and Tsipouras, Markos G and Tzimourta, Katerina D and Afrantou, Theodora and Ioannidis, Panagiotis and Tsalikakis, Dimitrios G and Sakkas, Konstantinos and Oikonomou, Emmanouil D and Grigoriadis, Nikolaos and others},
  journal={Data},
  volume={10},
  number={5},
  pages={64},
  year={2025},
  publisher={MDPI}
}

@article{dong2025self,
  title={Self-supervised spatial-temporal contrastive network for EEG-based brain network classification},
  author={Dong, Changxu and Sun, Dengdi and Luo, Bin},
  journal={Neural Networks},
  pages={107505},
  year={2025},
  publisher={Elsevier}
}

@inproceedings{khadka2024inducing,
  title={Inducing Inductive Bias in Vision Transformer for EEG Classification},
  author={Khadka, Rabindra and Lind, Pedro G and Mello, Gustavo and Riegler, Michael A and Yazidi, Anis},
  booktitle={ICASSP 2024-2024 IEEE International Conference on Acoustics, Speech and Signal Processing (ICASSP)},
  pages={2096--2100},
  year={2024},
  organization={IEEE}
}

@article{rafiei2022self,
  title={Self-supervised learning for electroencephalography},
  author={Rafiei, Mohammad H and Gauthier, Lynne V and Adeli, Hojjat and Takabi, Daniel},
  journal={IEEE Transactions on Neural Networks and Learning Systems},
  volume={35},
  number={2},
  pages={1457--1471},
  year={2022},
  publisher={IEEE}
}

@inproceedings{jiang2021self,
  title={Self-supervised contrastive learning for EEG-based sleep staging},
  author={Jiang, Xue and Zhao, Jianhui and Du, Bo and Yuan, Zhiyong},
  booktitle={2021 International Joint Conference on Neural Networks (IJCNN)},
  pages={1--8},
  year={2021},
  organization={IEEE}
}

@article{kostas2021bendr,
  title={BENDR: Using transformers and a contrastive self-supervised learning task to learn from massive amounts of EEG data},
  author={Kostas, Demetres and Aroca-Ouellette, Stephane and Rudzicz, Frank},
  journal={Frontiers in Human Neuroscience},
  volume={15},
  pages={653659},
  year={2021},
  publisher={Frontiers Media SA}
}

@article{krishnan2022self,
  title={Self-supervised learning in medicine and healthcare},
  author={Krishnan, Rayan and Rajpurkar, Pranav and Topol, Eric J},
  journal={Nature Biomedical Engineering},
  volume={6},
  number={12},
  pages={1346--1352},
  year={2022},
  publisher={Nature Publishing Group UK London}
}

@inproceedings{kumar2022muleeg,
  title={mulEEG: a multi-view representation learning on EEG signals},
  author={Kumar, Vamsi and Reddy, Likith and Kumar Sharma, Shivam and Dadi, Kamalaker and Yarra, Chiranjeevi and Bapi, Raju S and Rajendran, Srijithesh},
  booktitle={International Conference on Medical Image Computing and Computer-Assisted Intervention},
  pages={398--407},
  year={2022},
  organization={Springer}
}

@inproceedings{chen2020simple,
  title={A simple framework for contrastive learning of visual representations},
  author={Chen, Ting and Kornblith, Simon and Norouzi, Mohammad and Hinton, Geoffrey},
  booktitle={International conference on machine learning},
  pages={1597--1607},
  year={2020},
  organization={PMLR}
}

@article{foumani2024eeg2rep,
  title={Eeg2rep: enhancing self-supervised EEG representation through informative masked inputs},
  author={Foumani, Navid Mohammadi and Mackellar, Geoffrey and Ghane, Soheila and Irtza, Saad and Nguyen, Nam and Salehi, Mahsa},
  journal={arXiv preprint arXiv:2402.17772},
  year={2024}
}

@article{grill2020bootstrap,
  title={Bootstrap your own latent-a new approach to self-supervised learning},
  author={Grill, Jean-Bastien and Strub, Florian and Altch{\'e}, Florent and Tallec, Corentin and Richemond, Pierre and Buchatskaya, Elena and Doersch, Carl and Avila Pires, Bernardo and Guo, Zhaohan and Gheshlaghi Azar, Mohammad and others},
  journal={Advances in neural information processing systems},
  volume={33},
  pages={21271--21284},
  year={2020}
}

@inproceedings{zbontar2021barlow,
  title={Barlow twins: Self-supervised learning via redundancy reduction},
  author={Zbontar, Jure and Jing, Li and Misra, Ishan and LeCun, Yann and Deny, St{\'e}phane},
  booktitle={International conference on machine learning},
  pages={12310--12320},
  year={2021},
  organization={PMLR}
}

@inproceedings{he2022masked,
  title={Masked autoencoders are scalable vision learners},
  author={He, Kaiming and Chen, Xinlei and Xie, Saining and Li, Yanghao and Doll{\'a}r, Piotr and Girshick, Ross},
  booktitle={Proceedings of the IEEE/CVF conference on computer vision and pattern recognition},
  pages={16000--16009},
  year={2022}
}

@article{feichtenhofer2022masked,
  title={Masked autoencoders as spatiotemporal learners},
  author={Feichtenhofer, Christoph and Li, Yanghao and He, Kaiming and others},
  journal={Advances in neural information processing systems},
  volume={35},
  pages={35946--35958},
  year={2022}
}

@inproceedings{mohsenvand2020contrastive,
  title={Contrastive representation learning for electroencephalogram classification},
  author={Mohsenvand, Mostafa Neo and Izadi, Mohammad Rasool and Maes, Pattie},
  booktitle={Machine Learning for Health},
  pages={238--253},
  year={2020},
  organization={PMLR}
}

@misc{assran2022hidden,
      title={The Hidden Uniform Cluster Prior in Self-Supervised Learning}, 
      author={Mahmoud Assran and Randall Balestriero and Quentin Duval and Florian Bordes and Ishan Misra and Piotr Bojanowski and Pascal Vincent and Michael Rabbat and Nicolas Ballas},
      year={2022},
      eprint={2210.07277},
      archivePrefix={arXiv},
      primaryClass={cs.LG},
      url={https://arxiv.org/abs/2210.07277}, 
}

@inproceedings{li2022multi,
  title={A multi-view spectral-spatial-temporal masked autoencoder for decoding emotions with self-supervised learning},
  author={Li, Rui and Wang, Yiting and Zheng, Wei-Long and Lu, Bao-Liang},
  booktitle={Proceedings of the 30th ACM International Conference on Multimedia},
  pages={6--14},
  year={2022}
}

@inproceedings{assran2023self,
  title={Self-supervised learning from images with a joint-embedding predictive architecture},
  author={Assran, Mahmoud and Duval, Quentin and Misra, Ishan and Bojanowski, Piotr and Vincent, Pascal and Rabbat, Michael and LeCun, Yann and Ballas, Nicolas},
  booktitle={Proceedings of the IEEE/CVF Conference on Computer Vision and Pattern Recognition},
  pages={15619--15629},
  year={2023}
}

@article{guetschel2024s,
  title={S-JEPA: towards seamless cross-dataset transfer through dynamic spatial attention},
  author={Guetschel, Pierre and Moreau, Thomas and Tangermann, Michael},
  journal={arXiv preprint arXiv:2403.11772},
  year={2024}
}

@mastersthesis{hojjati2023multi,
  title={A Multi-View Self-Supervised Approach to Learn Representations of EEG Data for Downstream Prediction Tasks},
  author={Hojjati, Amirabbas},
  year={2023},
  school={NTNU}
}

@article{obeid2016temple,
  title={The temple university hospital EEG data corpus},
  author={Obeid, Iyad and Picone, Joseph},
  journal={Frontiers in neuroscience},
  volume={10},
  pages={196},
  year={2016},
  publisher={Frontiers Media SA}

}

@inproceedings{harati2014tuh,
  title={The TUH EEG CORPUS: A big data resource for automated EEG interpretation},
  author={Harati, Amir and Lopez, Silvia and Obeid, I and Picone, J and Jacobson, MP and Tobochnik, S},
  booktitle={2014 IEEE signal processing in medicine and biology symposium (SPMB)},
  pages={1--5},
  year={2014},
  organization={IEEE}
}

@article{khan2022nmt,
  title={The NMT scalp EEG dataset: an open-source annotated dataset of healthy and pathological EEG recordings for predictive modeling},
  author={Khan, Hassan Aqeel and Ul Ain, Rahat and Kamboh, Awais Mehmood and Butt, Hammad Tanveer and Shafait, Saima and Alamgir, Wasim and Stricker, Didier and Shafait, Faisal},
  journal={Frontiers in neuroscience},
  volume={15},
  pages={755817},
  year={2022},
  publisher={Frontiers Media SA}
}

@article{dosovitskiy2020image,
  title={An image is worth 16x16 words: Transformers for image recognition at scale},
  author={Dosovitskiy, Alexey and Beyer, Lucas and Kolesnikov, Alexander and Weissenborn, Dirk and Zhai, Xiaohua and Unterthiner, Thomas and Dehghani, Mostafa and Minderer, Matthias and Heigold, Georg and Gelly, Sylvain and others},
  journal={arXiv preprint arXiv:2010.11929},
  year={2020}
}

@article{loshchilov2017decoupled,
  title={Decoupled weight decay regularization},
  author={Loshchilov, Ilya and Hutter, Frank},
  journal={arXiv preprint arXiv:1711.05101},
  year={2017}
}

@article{sun2019multi,
  title={Multi-view intact space learning for tinnitus classification in resting state EEG},
  author={Sun, Zhi-Ran and Cai, Yue-Xin and Wang, Shao-Ju and Wang, Chang-Dong and Zheng, Yi-Qing and Chen, Yan-Hong and Chen, Yu-Chen},
  journal={Neural Processing Letters},
  volume={49},
  pages={611--624},
  year={2019},
  publisher={Springer}
}

@article{liu2019epileptic,
  title={Epileptic seizure prediction with multi-view convolutional neural networks},
  author={Liu, Chien-Liang and Xiao, Bin and Hsaio, Wen-Hoar and Tseng, Vincent S},
  journal={IEEE access},
  volume={7},
  pages={170352--170361},
  year={2019},
  publisher={IEEE}
}

@article{tang2020seizure,
  title={Seizure prediction using multi-view features and improved convolutional gated recurrent network},
  author={Tang, Lihan and Xie, Ning and Zhao, Menglian and Wu, Xiaobo},
  journal={IEEE Access},
  volume={8},
  pages={172352--172361},
  year={2020},
  publisher={IEEE}
}

@article{chen2022fusing,
  title={Fusing frequency-domain features and brain connectivity features for cross-subject emotion recognition},
  author={Chen, Chuangquan and Li, Zhencheng and Wan, Feng and Xu, Leicai and Bezerianos, Anastasios and Wang, Hongtao},
  journal={IEEE Transactions on Instrumentation and Measurement},
  volume={71},
  pages={1--15},
  year={2022},
  publisher={IEEE}
}

@article{xu2020recognition,
  title={Recognition of EEG signal motor imagery intention based on deep multi-view feature learning},
  author={Xu, Jiacan and Zheng, Hao and Wang, Jianhui and Li, Donglin and Fang, Xiaoke},
  journal={Sensors},
  volume={20},
  number={12},
  pages={3496},
  year={2020},
  publisher={MDPI}
}

@article{wan2023eegformer,
  title={EEGformer: A transformer--based brain activity classification method using EEG signal},
  author={Wan, Zhijiang and Li, Manyu and Liu, Shichang and Huang, Jiajin and Tan, Hai and Duan, Wenfeng},
  journal={Frontiers in Neuroscience},
  volume={17},
  pages={1148855},
  year={2023},
  publisher={Frontiers Media SA}
}

@article{chen2024eegformer,
  title={EEGFormer: Towards transferable and interpretable large-scale EEG foundation model},
  author={Chen, Yuqi and Ren, Kan and Song, Kaitao and Wang, Yansen and Wang, Yifan and Li, Dongsheng and Qiu, Lili},
  journal={arXiv preprint arXiv:2401.10278},
  year={2024}
}

@article{bardes2024revisiting,
  title={Revisiting feature prediction for learning visual representations from video},
  author={Bardes, Adrien and Garrido, Quentin and Ponce, Jean and Chen, Xinlei and Rabbat, Michael and LeCun, Yann and Assran, Mahmoud and Ballas, Nicolas},
  journal={arXiv preprint arXiv:2404.08471},
  year={2024}
}

@inproceedings{
dosovitskiy2021AnImage,
title={An Image is Worth 16x16 Words: Transformers for Image Recognition at Scale},
author={Alexey Dosovitskiy and Lucas Beyer and Alexander Kolesnikov and Dirk Weissenborn and Xiaohua Zhai and Thomas Unterthiner and Mostafa Dehghani and Matthias Minderer and Georg Heigold and Sylvain Gelly and Jakob Uszkoreit and Neil Houlsby},
booktitle={International Conference on Learning Representations},
year={2021},
url={https://openreview.net/forum?id=YicbFdNTTy}
}

@inproceedings{roy2019chrononet,
  title={ChronoNet: A deep recurrent neural network for abnormal EEG identification},
  author={Roy, Subhrajit and Kiral-Kornek, Isabell and Harrer, Stefan},
  booktitle={Artificial Intelligence in Medicine: 17th Conference on Artificial Intelligence in Medicine, AIME 2019, Poznan, Poland, June 26--29, 2019, Proceedings 17},
  pages={47--56},
  year={2019},
  organization={Springer}
}

@article{liu2021eeg,
  title={EEG power spectral analysis of abnormal cortical activations during REM/NREM sleep in obstructive sleep apnea},
  author={Liu, Shuling and Shen, Jiucheng and Li, Yezhou and Wang, Jing and Wang, Jianhua and Xu, Juan and Wang, Qiaojun and Chen, Rui},
  journal={Frontiers in neurology},
  volume={12},
  pages={643855},
  year={2021},
  publisher={Frontiers Media SA}
}

@inproceedings{Abnar2020QuantifyingAF,
  title={Quantifying Attention Flow in Transformers},
  author={Samira Abnar and Willem Zuidema},
  booktitle={Annual Meeting of the Association for Computational Linguistics},
  year={2020},
  url={https://api.semanticscholar.org/CorpusID:218487351}
}

@misc{mcinnes2020umapuniformmanifoldapproximation,
      title={UMAP: Uniform Manifold Approximation and Projection for Dimension Reduction}, 
      author={Leland McInnes and John Healy and James Melville},
      year={2020},
      eprint={1802.03426},
      archivePrefix={arXiv},
      primaryClass={stat.ML},
      url={https://arxiv.org/abs/1802.03426}
}
\end{document}